\documentclass{article}
\usepackage{iclr2027_conference,times}
\iclrfinalcopy
\usepackage[hidelinks]{hyperref}
\usepackage{url}
\usepackage{booktabs}
\usepackage{graphicx}
\usepackage{amsmath}
\usepackage{amssymb}
\usepackage{xcolor}

\newcommand{\readership}{\ensuremath{\rho}}

\title{What Looks Like a Capability Limit in\\Vision--Language Models Is a Readout Limit}

\author{Alfredo F.\ Frontera Del Valle\\
Columbia University\\
\texttt{aff2128@columbia.edu}}

\begin{document}
\maketitle
\lhead{Preprint. Under review.}
\thispagestyle{fancy}

\begin{abstract}
Benchmarks for vision--language models offer their answer choices in some
convention: a letter, a color name, a pixel coordinate. That convention is
treated as neutral. We find it is not, and that the limits a benchmark reports can
belong to the readout rather than to the model.

On 200 COCO photographs, Qwen3-VL-4B picks the correct one of nine locations for a
named object 68.5\% of the time when the locations are given in English and
20.0\% when the same nine locations are given as pixel coordinates. Chance is
11.1\%. The cost arises when the answer options are coordinates; giving the
model a coordinate in the question instead costs 3.5 points and is not
significant. The gap holds on a $4\times4$ grid, under 8-bit rather than 4-bit
quantization, and in every slice by object size, boundary distance and category. It also decides which model wins. Two models that
tie under English names differ by 39 points in one coordinate system and by 54 in
the other, in opposite directions.

On the color task, three of the four open models that can do the task at all show
the penalty; on photographs, two of three open models do, and so does Gemini, at
11.1 points on parseable answers ($p=10^{-4}$). GPT-4o does not. To ask whether a model reads a
coordinate at all, we attach the wrong name to each one and record which the model
follows. Color options written as hue angles are followed below chance; a
normalized pixel convention is followed at four times chance. This tells apart conventions a model can use from
ones it cannot, although it did not predict accuracy on two conventions we had not
tried. Five models also name the same color wheel five different ways, so a fixed
answer vocabulary is not neutral across models either.

Five times during this work we measured a capable model as incapable because our
scorer and the model disagreed about what an answer looks like. We report each
case. They are the phenomenon in miniature.
\end{abstract}

\section{Introduction}

Every benchmark for a vision--language model makes a choice it rarely discusses:
how the answer is to be written. A question about color may be answered with a
word or a hue angle; a question about position with a phrase, a grid reference, or
a pair of pixel coordinates. The benchmark author picks one, the score is computed
against it, and the number is reported as a property of the model. The convention
does not appear in the number.

We show that it should. On 200 photographs from COCO, Qwen3-VL-4B is asked where a
named object is and offered the nine cells of a $3\times3$ grid. When those nine
cells are described in English it answers correctly $68.5\%$ of the time. When the
same nine cells are described as pixel coordinates, in a prompt that also states
the image size and the axis conventions, it answers correctly $20.0\%$ of the time.
Chance is $11.1\%$. Nothing about the image, the object, the question, or the set
of alternatives has changed. The convention alone accounts for $48.5$ points.

The loss is not symmetric. If the model is instead handed a pixel coordinate and
asked which object sits there, the same convention costs $3.5$ points and is not
significant. A coordinate on the answer side is what costs the model; on the
question side it costs little. This matters because spatial benchmarks put
coordinates on the answer side, and a model that scores poorly on them may know
perfectly well where the object is.

The convention also decides comparisons between models. Qwen2.5-VL-7B and
Qwen3-VL-4B are statistically tied under English cell names, $68.0\%$ against
$68.5\%$. Under pixel coordinates the older model leads by $39$ points; under a
normalized coordinate convention the newer one leads by $54$. Both differences are
highly significant and they point in opposite directions. A leaderboard built on
either convention would rank these models confidently and, depending on the
convention, oppositely.

We then ask why a convention fails. Attaching an English name to each pixel
coordinate appears to repair the readout. Attaching the \emph{wrong} name shows
what actually happened: the model follows the name and abandons the coordinate,
and for hue angles it follows the coordinate less often than chance. The
coordinate was barely read. A normalized pixel convention, by contrast, is followed
four times more often than chance even against a contradicting name. We call the
share of responses that follow the coordinate, relative to chance, the
convention's \emph{readership}. It separates conventions a model can use from ones
it cannot. When we pre-registered a test of whether it also predicts accuracy on
two new conventions, it did not, and we report that.

Finally, the problem is not specific to coordinates. Asked to name the colors of
a calibrated wheel, five models produce five different stable vocabularies: only one has a red, only one a teal, and one admits just four names. A
fixed set of color words, the ordinary format of a color question, is therefore
not a neutral instrument across models either.

\paragraph{Contributions.}
\begin{itemize}
\item A convention swap that holds the image, the question, the alternatives, their
  order and the output format fixed and varies only how the alternatives are
  described, applied to color and to location, on synthetic stimuli and on real
  photographs (\S\ref{sec:color}, \S\ref{sec:space}).
\item Evidence that the convention changes not just scores but rankings, with two
  significant rank inversions among three open models on identical images
  (\S\ref{sec:models}).
\item A mispaired-name test that measures whether a convention is read at all,
  together with an honest account of what it does and does not predict
  (\S\ref{sec:readout}).
\item Replication on frontier models: the gap shrinks with capability and survives
  it in Gemini; it is absent in GPT-4o (\S\ref{sec:models}).
\item Per-model vocabulary calibration, and the finding that five models partition
  one color wheel five different ways (\S\ref{sec:calib}).
\item An audit of our own scoring in which we five times measured a capable model
  as incapable, reported as evidence for how easily this error is made
  (\S\ref{sec:audit}).
\end{itemize}

We do not claim that all benchmarks are wrong or that every coordinate system is
unreadable, and where we attribute a convention to a model's training we cite the
model's own report. We claim that the answer convention is a variable, that it is
often large, and that it is currently invisible.

\section{Related work}
\label{sec:related}

\paragraph{The answer interface.} The closest prior work is
\citet{provise2026}, which names the problem in nearly our terms: spatial
benchmarks demand coordinates, options or text, and call the resulting gap an
answer-interface mismatch. Their remedy is to change the output modality, letting
image-generation models draw their answers and parsing the drawings back into the
original metrics. We keep the text interface fixed and change only how the
alternatives are described, which lets us compare conventions on identical items,
and we add the mispaired-name test to ask whether a convention is read at all.
Neither the paired comparison nor the mispaired-name test appears in that work.

\paragraph{Benchmarks that are too easy.} \citet{seeingwithoutlooking2026} and
\citet{mmgist2026} show that many benchmark items can be answered without the
image, from leakage in the question text or from world knowledge. Our finding runs
the other way: a benchmark can be too hard for a reason that has nothing to do
with the model's vision. The two distortions are not exclusive. A single reported
number can be inflated by leakage and deflated by convention at the same time, and
neither is visible in the number. \citet{seeorguess2026} varies the phrasing of
the question to measure reliance on textual priors; we hold the question fixed and
vary the description of the answers.

\paragraph{Selection bias in multiple choice.} A well-developed line of work
shows that language models and vision--language models prefer particular option
positions and particular option tokens, and mitigates this by permuting order
\citep{mmbench2024}, debiasing the prior over labels \citep{pride2024}, or
separating positional from token effects \citep{lvlmselectionbias2025}. Every
method in this line holds the option \emph{descriptions} fixed. In our design that
corresponds to the single rung in which a color word is replaced by an integer
label pointing at the same word, and that rung costs one point. The thirty-point
drop occurs one step earlier, when the description itself changes, and selection
bias methods do not reach it. \citet{prefill2025} show that first-token scoring
can be made reliable by prefilling the answer prefix; our audit reaches the same
conclusion from the failure side, by refusing to score any alphabet the model does
not place mass on.

\paragraph{Grounding models.} Systems such as \citet{glamm2024} and benchmarks
such as \citet{pointarena2025} treat coordinate output as a capability to build and
to measure. We do not disagree. We measure coordinate output as an instrument that
can misreport a capability the model has, and our framework predicts that a model
trained to emit coordinates will show a smaller convention gap. A separate line
treats the textual coordinate readout itself as the defect and replaces or
retrains it: explicit position-to-coordinate tokens \citep{ruler2025},
attention-based pointing with no coordinate tokens at all \citep{guiactor2025},
parallel box decoding \citep{locateanything2026}, and distillation against
geometric rather than token loss \citep{ioupd2026}. Those are remedies on the
model side; ours is the measurement they presuppose, of how much a fixed model's
score depends on the readout it is handed.

\paragraph{Color stimuli.} The wheel is the CIELAB circle at fixed lightness
standard in human visual working memory research
\citep{zhangluck2008,bayshusain2008}, used for its perceptual uniformity; the
paper makes no comparison to people.

\section{Method}

Every experiment holds the image, the question, the candidate set, the candidate
order and the required output format fixed, and varies only how the candidates are
\emph{described}. We call that manipulation a \emph{convention swap}. Three further
components make it interpretable: a response vocabulary calibrated per model, a
capability gate, and a held-out split declared before the data existed.

\subsection{Calibrating the response vocabulary}
\label{sec:calib}

A convention swap needs a control condition the model can certainly express. We do
not assume one. For each model we render one swatch per wheel angle on a
$5^\circ$ grid, three times per angle under different square sizes and placements,
and ask it to name the color in free report. An angle is retained only if all
three renders elicit the same word. Anchors are the centers of contiguous
unanimously named runs, separated by at least $40^\circ$.

Calibration yield is an outcome, not a setting. The five models tested here do not
agree. Qwen3-VL-4B's wheel contains no red: the swatch a coordinate table calls
red renders \texttt{(238,146,146)}, and the model calls it pink, correctly.
SmolVLM-2B's wheel does have a red. Qwen2.5-VL-7B has a stable teal where
Qwen3-VL has none. A fixed
textual answer set therefore need not carry equivalent perceptual grounding across
models (Table~\ref{tab:models}, Appendix~\ref{app:tables}).

We also resolve the \emph{surface} of each candidate token per model rather than
assuming it. Candidate ids are taken from whichever surface variant (bare, leading space,
capitalized, and their combinations) places the most probability mass on the
candidate set under probes with known answers. A model whose candidate mass never
clears $0.5$ has no restricted first-token readout for that alphabet and is refused
rather than reported. Section~\ref{sec:audit} explains why this guard exists.

\subsection{The convention swap}

Within a trial the model chooses among $K$ alternatives that denote the same $K$
referents in every condition. Only their surface description changes: a color word
versus a hue angle, an English spatial phrase versus a pixel pair. Output format is
held fixed within a comparison, so the cost of emitting a digit instead of a word
is measured separately and does not contribute; it is $1$ point
(Table~\ref{tab:color}).

\subsection{Visible and cached conditions, and the cue crossover}
\label{sec:crossover}

For the color experiments each trial is run two ways. In the \emph{visible}
condition the array is in the prompt with the question. In the \emph{cached}
condition the array is encoded once into the model's key--value cache with an
acknowledgement turn, and the question is a text-only continuation with no image
present; the model answers from what it retained. Both are scored the same way.

To check that a cached answer depends on the remembered binding at the cued
location, and not on the array as a whole, we also run a \emph{cue crossover}.
Two arrays differ only by swapping the colors at two locations. Asking about the
first location should shift the score contrast one way, asking about the second
should shift it the other way, and asking about a third, untouched location should
not shift it at all. The reported \emph{interaction} is the difference between the
first two shifts, normalized by the spread of the candidate scores so that
conditions with different score ranges can be compared. A location-blind account
predicts no sign reversal.

\subsection{The mispaired gloss, and readership}
\label{sec:readership}

Attaching a name to a coordinate, as in \texttt{1=(66,50) [top-left]}, raises
accuracy, but accuracy alone cannot say whether the coordinate was used or merely
accompanied.
We therefore \emph{derange} the gloss so that no option carries its true pairing.
Every coordinate and every name still appears exactly once; only the pairing breaks,
and the derangement is verified to have no fixed point, so the two resulting keys
can never coincide. One response is then scored against both:
\textbf{key\_coord}, the option whose coordinate is correct, and \textbf{key\_name},
the option whose glossed name is correct.

We define \emph{readership} as
\begin{equation}
\readership \;=\; \frac{P(\text{response}=\text{key\_coord} \mid \text{keys disagree})}{1/K},
\end{equation}
the share of responses following the coordinate, relative to chance. $\readership
\approx 1$ means the coordinate carries no information the model uses; $\readership
< 1$ means it is actively avoided, since the correct coordinate now sits beside the
wrong name.

\subsection{Capability gate}

A convention gap is a difference between two readouts of the same competence. Where
there is no competence, a null contrast is uninformative rather than disconfirming.
We therefore require the control condition to exceed chance by $20$ points
\emph{on the readout the contrast is computed with}. Gating on a different readout
than the contrast uses admits models whose contrast is measured by an instrument
with no signal in its own baseline; we made exactly that error and corrected it.

\subsection{Held-out split}

Because the vocabulary is chosen by asking the model, evaluating at the same angles
cannot separate ``the model has this representation'' from ``we selected the colors
it likes''. The held-out rule was committed before the generator was written:
integer angles $10$--$20^\circ$ from each anchor, off the calibration grid, and at
least $20^\circ$ from every other anchor. All hypotheses, endpoints, exclusions and
failure conditions in this paper were registered in a versioned analysis plan before
the corresponding code existed; two of our own predictions were refuted by it and
are reported as such (Sections~\ref{sec:color} and~\ref{sec:readout}).

\section{Color: the convention gradient}
\label{sec:color}

We first establish the effect where every factor can be controlled. Stimuli are
four colored squares on a gray field, drawn from a vocabulary calibrated per
model (\S\ref{sec:calib}); the model is asked which color occupied a cued
location. All figures below are on held-out angles, never used to select the
vocabulary, sampled by a rule committed before the generator was written.

\begin{table}[t]
\centering
\caption{Color, Qwen3-VL-4B, 173 held-out trials. Only the description of the six
alternatives changes; stimulus, candidate set, candidate order and output format
are fixed. Interaction is the spread-normalized cue-crossover statistic (\S\ref{sec:crossover}); chance is $16.7\%$.}
\label{tab:color}
\begin{tabular}{lrrrr}
\toprule
options described as & visible & cached & interaction & \% of word \\
\midrule
color word                      & 84.4 & 85.0 & 9.27 & 100 \\
integer label, word listed       & 83.8 & 83.5 & 8.99 & 97 \\
hue angle with the word attached & 77.5 & 80.1 & 8.22 & 89 \\
hue angle, legend stated once    & 53.8 & 63.2 & 6.34 & 68 \\
hue angle alone                  & 21.8 & 22.3 & 1.08 & 12 \\
\bottomrule
\end{tabular}
\end{table}

Three facts in Table~\ref{tab:color} matter. Changing the \emph{output} from a
color word to an integer costs $1$ point, so what follows is not about emitting
digits. Changing the \emph{description} of the same six alternatives from words to
hue angles costs $30.1$ points visible ($175$ label-only correct against $19$
numeric-only, exact McNemar $p=8.7\times10^{-33}$, $n=519$). And a hue angle with
no grounding at all sits at $21.8\%$ against $16.7\%$ chance: it is not a degraded
readout, it is not a readout.

Accuracy and the cue-crossover interaction move together across all five rows. The
crossover swaps two remembered colors between two cued locations and predicts a
sign reversal that a location-blind account cannot produce; a third, untouched
location is null at every row ($-0.55$ to $+0.34$). That the evidence measure
tracks accuracy rules out an account in which only response formatting changes.
The cached column shows the same gradient with no image in context: the
convention, not the availability of the pixels, sets the score. On the
calibration angles themselves the word conditions score near $99\%$; the drop to
$84\%$ on held-out angles is the price of the split, and the reason it exists. No
benchmark scores color in CIELAB hue angles; the hue rows are the limiting case of
a convention with no grounding at all.

\paragraph{Grounding is not what the gloss supplies.} Attaching the word to the
angle appears to repair the readout, and does not. Deranging the gloss so that no
option carries its true pairing separates the two keys: the model follows the
\emph{name} on $75.1\%$ of cells and the \emph{angle} on $5.1\%$, against $16.7\%$
chance ($p=7.7\times10^{-21}$ below chance, $n=692$). The correct angle is not
ignored but actively avoided, because it now sits beside the wrong word. The
apparent recovery is substitution.

\section{Space, and real photographs}
\label{sec:space}

If the deficit were about color it would stop here. We repeat the manipulation on
location, first on synthetic arrays and then on COCO val2017 photographs \citep{coco2014}. On the
synthetic arrays, four colored squares in nine cells, the locate ordering is cell
name $70.8\%$, normalized $58.3\%$, pixel $42.5\%$, and the same with the array
cached (Table~\ref{tab:synthspace}, Appendix~\ref{app:tables}). On COCO, one
object per image, its position asked for as a choice among the nine cells of a
$3\times3$ grid; only the description of those nine cells varies. Admission
criteria were fixed before the data was touched: a unique instance of the
category, segmentation area under $25\%$ of the image, and a center at least $8\%$
of the image inside its cell's border.

\begin{table}[t]
\centering
\caption{COCO val2017, 200 photographs. \textsc{locate} asks for the position of a
named object (9 options, chance $11.1\%$); \textsc{identify} supplies the position
and asks which object is there (6 options, chance $16.7\%$). $\Delta$ is against
English cell names, exact McNemar; ${}^{***}$ $p<10^{-3}$. The deranged row is
scored on the coordinate key.}
\label{tab:coco}
\begin{tabular}{lrrrr}
\toprule
& \multicolumn{2}{c}{\textsc{locate}} & \multicolumn{2}{c}{\textsc{identify}} \\
\cmidrule(lr){2-3}\cmidrule(lr){4-5}
options described as & acc. & $\Delta$ & acc. & $\Delta$ \\
\midrule
English cell name       & 68.5 & ---            & 96.0 & --- \\
normalized 0--999       & 59.5 & $-9.0$         & 97.0 & $+1.0$ \\
absolute pixels         & 20.0 & $-48.5^{***}$  & 92.5 & $-3.5$ \\
pixels, name attached   & 39.0 & $-29.5^{***}$  & 95.5 & $-0.5$ \\
pixels, name deranged   & 16.0 & $-52.5^{***}$  & 93.5 & $-2.5$ \\
\bottomrule
\end{tabular}
\end{table}

\paragraph{A coordinate on the answer side is costly; on the question side it is
almost free.} Table~\ref{tab:coco} is the paper's central result. Asking where an
object is, with the nine candidates written in pixels, scores $20.0\%$ against
$68.5\%$ in English, a drop of $48.5$ points ($95\%$ bootstrap interval $[-56.5,
-40.0]$, $p=1.2\times10^{-21}$) on a task whose chance rate is $11.1\%$. Handing
the model the same pixel coordinate as \emph{context} and asking what is there
costs $3.5$ points and is not significant ($p=0.065$). The asymmetry replicates in
InternVL3-2B ($-32.5$, $p=4.2\times10^{-13}$ on the answer side; $-8.0$ on the
question side). On a $4\times4$ grid with sixteen cells and $6.25\%$ chance, cell
names score $26.0\%$ and pixels $7.5\%$ ($p=3.0\times10^{-6}$): at finer
granularity pixels fall to chance while names stay four times above it. The gap
holds within every tercile of object size and of distance from a cell boundary, and
in eleven of twelve COCO categories with at least five images
(Appendix~\ref{app:tables}). The pixel prompt states the original image size;
the vision encoders rescale by a few percent at most, less than the $8\%$ margin
between a cell center and its border, so the frame cannot move an answer across a
cell.

\paragraph{Prompt length is ruled out from inside the design.} In the synthetic
spatial task the normalized-coordinate prompt is \emph{longer} than the pixel
prompt, 187 tokens against 183, and scores 15.8 points higher. No external
control is required.

\paragraph{The readable convention is model-specific.} The normalized $0$--$999$
convention nearly rescues Qwen3-VL ($59.5\%$ against $20.0\%$ for pixels) and
barely helps InternVL3-2B ($19.0\%$ against $13.5\%$). Qwen3-VL's technical
report describes a normalized $[0,1000]$ grounding system \citep{qwen3vl2025};
the advantage of that convention does not transfer to a model trained differently.
This mirrors what calibration already showed for color: models do not share an
answer vocabulary, and they do not share a readable coordinate system either.

\section{Readership: is the convention read at all?}
\label{sec:readout}

Accuracy under a convention conflates two things: whether the model can use the
convention and whether it happens to reach the right answer by some other route.
The mispaired-name test of \S\ref{sec:readership} separates them. Each coordinate
is glossed with a name that belongs to a different option, and one response is
scored against both keys.

\begin{table}[t]
\centering
\caption{Which key does the model follow when coordinate and name disagree?
Readership is the share following the coordinate, relative to chance. Qwen3-VL-4B
throughout; $n$ is scored cells.}
\label{tab:readership}
\begin{tabular}{lrrrrr}
\toprule
convention & follows coord. & follows name & neither & readership & $n$ \\
\midrule
CIELAB hue, name inline   &  5.1\% & 75.1\% & 19.8\% & $0.30\times$ & 692 \\
CIELAB hue, legend        &  8.2\% & 61.7\% & 30.1\% & $0.49\times$ & 692 \\
pixels, synthetic grid    & 21.7\% & 31.7\% & 46.7\% & $1.95\times$ & 120 \\
pixels, COCO              & 16.0\% & 25.0\% & 59.0\% & $1.44\times$ & 200 \\
normalized 0--999, COCO   & 45.0\% & 14.0\% & 41.0\% & $4.05\times$ & 200 \\
\bottomrule
\end{tabular}
\end{table}

\paragraph{Hue angles are avoided, not ignored.} With a mispaired name beside it,
the correct hue angle is chosen on $5.1\%$ of cells against a chance rate of
$16.7\%$ ($p=7.7\times10^{-21}$ below chance). The model is not indifferent to the
coordinate; it moves away from the slot carrying the correct angle because that
slot now carries the wrong word. The same holds when the pairing is stated once in
a legend rather than inline. What looked in Table~\ref{tab:color} like a gloss
repairing the readout was the gloss replacing it.

\paragraph{A familiar convention is read even against a contradiction.} The
normalized $0$--$999$ convention is followed on $45.0\%$ of mispaired trials
against $11.1\%$ chance, and the name on only $14.0\%$. The preference reverses
completely between the two ends of the table. Nothing differs between those rows
except the units, and the model's behavior with the units when they disagree with
the words is the cleanest evidence we have that some conventions are read and
others are not.

\paragraph{What readership does not do.} Before running two further conventions
we registered the prediction that accuracy would be monotone in readership and
placed each new convention relative to the pixel and normalized anchors. A grid
reference ($A1$ to $C3$) had readership $1.26\times$ and a percentage convention
$1.30\times$, both below pixels, so both were predicted to score under $20.0\%$.
They scored $31.0\%$ and $23.0\%$, in the wrong order; the prediction failed on
every part. The reason is visible in the table: readership mixes how usable a
convention is with how hard a competing English name pulls, and the grid reference has the strongest name pull of any convention tested on photographs. We therefore report readership as
a test of whether a convention is read, which it is, and not as a predictor of
accuracy, which it is not, and we did not search for a statistic that fits the
five points we have.

Readership for each model on COCO pixels is given in
Appendix~\ref{app:extra}.

\section{Across models}
\label{sec:models}

\paragraph{Five vocabularies for one wheel.} Each model was calibrated separately
by the procedure of \S\ref{sec:calib}. Table~\ref{tab:models}
(Appendix~\ref{app:tables}) shows what they returned. Green near $145^\circ$ is the only anchor all five share. Qwen3-VL's
wheel has no red; SmolVLM's does. Qwen2.5-VL stabilizes a teal that Qwen3-VL,
one generation later in the same family, cannot. LLaVA-1.5 admits four stable
names where the others admit six. A color question with a fixed set of answer
words is asking different models different questions.

\paragraph{The color gap across models.} On held-out angles with each model's own
vocabulary, the word-to-numeric penalty is $30.1$ points in Qwen3-VL-4B
($p=8.7\times10^{-33}$), $23.9$ in Qwen2.5-VL-7B ($p=9.1\times10^{-4}$) and
$19.7$ in InternVL3-2B ($p=1.6\times10^{-2}$). LLaVA-1.5-7B clears the capability
gate and shows no penalty ($-2.1$, $p=1.0$); it is the oldest model tested and the
one with the fewest stable colors, which we report as description and not as
explanation. SmolVLM-2B does not clear the gate and is uninformative. InternVL3
shares a Qwen language backbone, so it is a partial cross-family replication and
we describe it as one.

\begin{figure}[t]
\centering
\includegraphics[width=\linewidth]{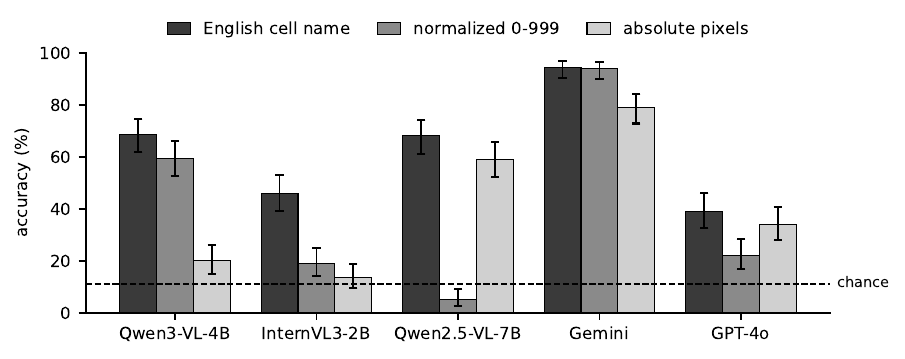}
\caption{COCO \textsc{locate} on the same 200 photographs; chance is $11.1\%$
(dashed). Bars are accuracy with $95\%$ Wilson intervals. The cell-name minus pixel
gap, paired per image with a $95\%$ bootstrap interval over images, is $-48.5$
$[-56.5,-40.0]$ for Qwen3-VL-4B, $-32.5$ $[-40.5,-24.5]$ for InternVL3-2B,
$-9.0$ $[-18.5,0.0]$ for Qwen2.5-VL-7B, $-15.5$ $[-21.5,-10.0]$ for Gemini and
$-5.0$ $[-12.5,2.5]$ for GPT-4o. Exact values, the 8-bit run and the
parseable-only analysis are in Table~\ref{tab:fivecoco},
Appendix~\ref{app:tables}.}
\label{fig:coco}
\end{figure}

\paragraph{Real photographs, five models.} Figure~\ref{fig:coco} gives the
locate task on the same 200 COCO images for three open models run locally and two
frontier models run through their APIs (\texttt{gemini-flash-latest} and
\texttt{gpt-4o} as served on 20 September 2026). All saw byte-identical prompts.
LLaVA-1.5 and SmolVLM are absent because the harness refused to score them: no
first-token digit surface carried probability mass (\S\ref{sec:audit}). The
frontier models are scored by generate-and-parse because their APIs expose no
usable first-token distribution; the local models are scored by first token and,
where both are available, the two readouts agree on 97--100\% of trials (Appendix~\ref{app:prompts}).
Every headline $p$-value is below $10^{-3}$ except InternVL3-2B's color gap.

\paragraph{Quantization makes the gap larger, not smaller.} Running Qwen3-VL-4B at
8-bit rather than 4-bit raises every condition and widens the gap from $48.5$ to
$52.0$ points. The effect is not an artifact of low precision.

\paragraph{The gap survives at the frontier, and shrinks.} Gemini scores $94.5\%$
under English names and still loses $11.1$ points to pixel coordinates on
parseable answers. GPT-4o, which is weak at this task under every convention,
shows no significant gap. The convention penalty is largest in small open models,
smaller in a frontier model, and present in one of the two frontier models we
could test.

\paragraph{Rankings reverse.} Table~\ref{tab:inversion} pairs the models on the
same images. Qwen2.5-VL-7B and Qwen3-VL-4B are tied under English names. Under
pixels the older model leads by $39$ points; under the normalized convention the
newer one leads by $54$. InternVL3-2B trails Qwen2.5-VL under names and pixels
and leads it under the normalized convention. Two of three pairs invert. We
registered the search for inversions before running the models and report every
pair, inverting or not. The direction of the Qwen inversion follows the two
models' documented training conventions: Qwen2.5-VL's report states that it
represents boxes and points in coordinates based on the actual dimensions of the
input image \citep{qwen25vl2025}, and Qwen3-VL's describes a normalized
$[0,1000]$ system \citep{qwen3vl2025}. Each model leads in the convention its
own report says it was trained in.

\begin{table}[t]
\centering
\caption{Paired differences on the same 200 photographs, exact McNemar. A sign
change across rows within a pair is a rank inversion.}
\label{tab:inversion}
\begin{tabular}{llrr}
\toprule
pair & convention & difference & $p$ \\
\midrule
Qwen2.5-VL-7B vs Qwen3-VL-4B & cell name   & $-0.5$  & $1.0$ \\
                             & pixel       & $+39.0$ & $8.6\times10^{-14}$ \\
                             & normalized  & $-54.5$ & $1.2\times10^{-30}$ \\
\midrule
InternVL3-2B vs Qwen2.5-VL-7B & cell name  & $-22.0$ & $5.7\times10^{-7}$ \\
                              & pixel      & $-45.5$ & $8.5\times10^{-19}$ \\
                              & normalized & $+14.0$ & $1.5\times10^{-5}$ \\
\midrule
InternVL3-2B vs Qwen3-VL-4B  & cell name   & $-22.5$ & $3.6\times10^{-7}$ \\
                             & pixel       & $-6.5$  & $0.07$ \\
                             & normalized  & $-40.5$ & $1.9\times10^{-22}$ \\
\bottomrule
\end{tabular}
\end{table}

Qwen2.5-VL-7B's $5.0\%$ under the normalized convention is below chance for the
same reason: on $53.0\%$ of trials it chooses the option whose $0$--$999$ numbers,
read as pixels, lie nearest the object (null $11.8\%$, $p=7.8\times10^{-48}$), and
that option is the true target only $2.5\%$ of the time. Qwen3-VL-4B shows the
mirror on pixels, weaker; InternVL3-2B shows neither. The test was post hoc
(Appendix~\ref{app:extra}).

\section{A readout audit of our own experiments}
\label{sec:audit}

Five times in this work we scored a model as performing at or near chance, looked
at what it had actually produced, and found it answering correctly. In each case
the scorer and the model disagreed about what an answer looks like. We report
them because they are the paper's claim at the scale of a single line of code,
and because each one was invisible from the score.

\paragraph{The wrong token surface.} SmolVLM-2B scored $16$--$30\%$ against
$16.7\%$ chance on every color condition, with candidate mass $0.00$. It was
emitting \texttt{~Green}, leading space and capital, token 4910; the scorer read
\texttt{green}, token 9298. Its top three first tokens on a green swatch were
\texttt{~Green}, \texttt{~Red}, \texttt{~Blue}, the correct one far ahead. The fix
is to resolve each model's emitted surface by measurement, and to refuse to report
when no surface carries mass.

\paragraph{No first-token readout exists.} For SmolVLM and LLaVA-1.5 every
integer label begins with the same shared leading-space token. A restricted
first-token score over such labels is not noisy; it does not exist. The harness
now refuses rather than emitting chance-level numbers, and those models are
scored by generating and parsing.

\paragraph{Held-out stimuli in a discarded category.} The held-out rule kept
color angles $20^\circ$ from every retained anchor but not from names the
calibration had discarded. Seven of 24 held-out angles sat where the model says
brown, green or blue while the key said yellow, orange or purple. The model was
right; the key was wrong. The defect was conservative, penalizing the word
condition at least as much as the numeric one, and the registered analysis
stands; the rule was still wrong and the amendment says so.

\paragraph{Two index spaces.} The cross-model parser returned an index into the
shuffled option list while the target indexed the canonical alphabet. Every
cross-model word score sat at chance, four models at once, and it looked like
four weak models. Only the word rung was affected, because the label rungs used
slot indices on both sides, which is exactly why it looked like a property of the
models. After the fix the two readouts agree on 97--100\% of trials.

\paragraph{Output truncated before the answer.} The first frontier run returned
empty strings on ten of ten calls: the model spends its output budget on
reasoning tokens before the answer, and the budget was eight tokens. A capable
model would have been recorded as producing nothing. It was caught only because
the raw text was printed beside the score.

None was found by looking at the score; all were found by looking at what the
model said. The repository's audits now parse the prompt against the target,
recompute every label from raw output, break the trial-to-answer pairing and
require chance, and refuse alphabets the model does not emit.

\section{Limitations}

The real-image results come from one dataset, COCO, with categories skewed toward
\emph{person}, and one task: choosing among described grid cells for a named
object. That is a selection task, a stand-in for how multiple-choice spatial
benchmarks are written, not a grounding benchmark with free-form box output, and
nothing here licenses a claim about box regression. A $3\times3$ and a
$4\times4$ grid are coarse. The color results use
synthetic squares on a gray field. Two frontier models were tested and only one
shows the effect; a third might go either way. InternVL3-2B shares a Qwen language
backbone, so the cross-family replication is partial. The frontier models are
scored by generating and parsing while the local models are scored by first
token; the two agree wherever both are available, but they are not the same
instrument. The primary model runs at 4-bit precision, and although 8-bit widens
the gap, full precision was not run. All decoding is greedy, which is standard for
evaluation and is what the benchmarks we compare against use, but sampling was
not tested. The prompts state each convention once, in one wording; whether a
longer instruction that spells out the mapping narrows the gap was not tested. And readership, the quantity we
introduce, is a valid test of whether a convention is read and a failed predictor
of accuracy; the pre-registered out-of-sample test missed on every part.

\section{What a benchmark author should do}

Report the convention with the number; a localization score without the
coordinate system it was scored in is not interpretable across papers. Run the
swap: describe the same alternatives a second way, in plain language, and report
both, because a difference of more than a few points is a property of the
instrument and belongs next to the result. Run the mispaired test before trusting
a coordinate readout; a model that follows the coordinate at chance or below
against a wrong name is not reading it. Do not gloss coordinates to help, since
that turns the task into a naming task and the number stops being about
coordinates. Separate parse failures from wrong answers, because a score that
counts both as wrong hides which is happening. Calibrate the answer vocabulary per
model; a fixed word list is a choice about which model the benchmark is easiest
for. And check the scorer against the raw output: every error in
\S\ref{sec:audit} was invisible from the score and obvious from the text.

\section*{Ethics statement}

The work uses a public dataset (COCO) and synthetic stimuli, involves no human
subjects or personal data, and its intended effect is more accurate evaluation of
publicly available models.

\section*{Reproducibility statement}

Every number and figure in this paper is produced by scripts from the raw per-trial logs, none entered by hand, and a check script confirms that every decimal in this manuscript appears in their output. Two audit scripts verify, among other
things, that the option at each stored target index actually names the target when
the prompt string itself is parsed, that every stored label recomputes from the raw
model output, that breaking the trial-to-answer pairing drives accuracy to chance,
and that no alphabet is scored that the model does not emit. All hypotheses,
endpoints, exclusions, failure conditions and amendments are recorded with
timestamps in a versioned analysis plan, including the two predictions that failed. Code, logs, prompts, the analysis plan and both audits are available from the author on request and will be released publicly.

\section*{Use of AI assistance}

The author conceived the project and its
experimental program, set the research question, and decided the direction at
every stage: what to test, what to read, which framings to abandon when their
evidence did not hold, which intermediate results to reject as insufficient,
which outside critiques to act on, and what to claim and submit. The author
specified each experiment and each analysis, directed the literature search and
chose what to read from it, and required that every claim survive a plain-language
explanation and an independent audit before it was accepted. The labor-intensive
parts of the work, principally writing code to the author's specification, running
batches of trials, retrieving candidate references, and producing first drafts of
text and tables, were delegated to AI tools, mainly Claude via Claude Code, with
ChatGPT and Perplexity used to obtain independent critiques of the work in
progress. The author implemented the experimental program on personal hardware and
accounts, reviewed the output of every delegated step, and revised the text
throughout; the final wording, structure and claims are the author's. Five scoring
errors described in Section~\ref{sec:audit} arose during AI-assisted
implementation and were caught by the audits the author required. Every number in the paper is regenerated from raw logs by scripts in the repository, and the
pre-registered hypotheses, including the two that failed, are recorded with
timestamps in the same repository.

\bibliographystyle{iclr2027_conference}
\bibliography{refs}

\appendix
\section{Additional observations}
\label{app:extra}

\paragraph{Readership across models.} On COCO pixels, readership is $1.12\times$
for InternVL3-2B, $1.44\times$ for Qwen3-VL-4B, $1.89\times$ for GPT-4o,
$4.86\times$ for Qwen2.5-VL-7B and $6.62\times$ for Gemini. Normalizing each
model's convention gap by its headroom (cell-name accuracy minus chance), the two
models that barely read pixels lose $85$--$93\%$ of that headroom to the
convention and the three that read them lose $16$--$19\%$. With five points this
is an observation, not a fit, and it was not pre-registered.

\paragraph{The number below chance.} Qwen2.5-VL-7B scores $5.0\%$ on the
normalized convention, below chance and below the $10.9\%$ its own answer bias
would produce against the target distribution. We first tested whether it applies
a systematic transform to the grid: the identity, transpose, anti-transpose, both
flips, three rotations, and row and column shifts. None explains it; the
anti-transpose reaches $19\%$ ($p=7\times10^{-4}$) and accounts for under a fifth
of responses. The eleventh hypothesis, prompted by an outside review after those
ten had failed, is that the model reads the $0$--$999$ numbers in its training
units, as pixels. For each trial we take the option whose numbers, read as pixels,
lie nearest the object's box center and ask whether the model chose it. It did on
$53.0\%$ of trials, against a null of $11.8\%$ from the product of the model's
marginal answer distribution and the predicted-slot distribution ($p=7.8\times
10^{-48}$, exact binomial against $1/9$). That option is the true target on only
$2.5\%$ of trials, which is why the score falls below chance rather than to it.
The same test applied to Qwen3-VL-4B on pixel numbers, read as thousandths of the
image, gives $28.0\%$ against $11.4\%$ ($p=4.6\times10^{-11}$); applied to
InternVL3-2B it gives $10.5\%$ against $10.2\%$ ($p=0.64$). An accurate model
can inflate this rate through trials where the misread option happens to be the
target; excluding those leaves $54.4\%$ for Qwen2.5-VL and $26.4\%$ for Qwen3-VL.
Applied to Gemini and GPT-4o in both directions, the test is null ($0.0$ to
$15.0\%$ against nulls of $10.9$ to $11.5\%$, smallest $p=0.07$): the misreading
accounts for the open-model inversions and not for the frontier gaps. The script
and its output are in the repository, and the protocol file records the test as
post hoc.

\section{Prompt templates, option order, gate sensitivity, and instrument agreement}
\label{app:prompts}

\paragraph{Real photographs.} Every prompt is one turn of text under the image.
The examples below are the verbatim prompts for one image (COCO 139, $640\times426$),
taken from the logs. The nine cells are listed in an order fixed per image and
identical across conventions.

{\small\ttfamily\raggedright
\textbf{cell name, locate.} In this image, the microwave is at one of these
positions: 1=center, 2=bottom-right, 3=middle-left, 4=top-right, 5=top-center,
6=middle-right, 7=bottom-center, 8=top-left, 9=bottom-left. Which one? Answer
with a single integer 1 to 9 and nothing else.

\textbf{absolute pixels, locate.} Positions are given as (x, y) pixel
coordinates in this 640 by 426 image, where (0, 0) is the top-left corner and x
increases to the right and y increases downward. In this image, the microwave is
at one of these positions: 1=(320, 213), 2=(533, 355), 3=(106, 213), 4=(533, 71),
5=(320, 71), 6=(533, 213), 7=(320, 355), 8=(106, 71), 9=(106, 355). Which one?
Answer with a single integer 1 to 9 and nothing else.

\textbf{normalized, locate.} Positions are given as (x, y) coordinates
normalized to a 0 to 999 range, where (0, 0) is the top-left corner and x
increases to the right and y increases downward. In this image, the microwave is
at one of these positions: 1=(500, 500), 2=(833, 833), 3=(166, 500),
4=(833, 167), 5=(500, 167), 6=(833, 500), 7=(500, 833), 8=(166, 167),
9=(166, 833). Which one? Answer with a single integer 1 to 9 and nothing else.

\textbf{glossed pixels, locate.} [same preamble] \ldots{} 1=(320, 213) [center],
2=(533, 355) [bottom-right], 3=(106, 213) [middle-left], \ldots

\textbf{mispaired pixels, locate.} [same preamble] \ldots{} 1=(320, 213)
[middle-left], 2=(533, 355) [top-right], 3=(106, 213) [top-left], \ldots

\textbf{cell name, identify (coordinate on the question side).} In this
image, what object is at position middle-right? It is one of these: 1=fork,
2=kite, 3=bottle, 4=microwave, 5=baseball glove, 6=donut. Answer with a single
integer 1 to 6 and nothing else.

\textbf{absolute pixels, identify.} [same preamble] In this image, what object
is at position (533, 213)? It is one of these: 1=fork, 2=kite, 3=bottle,
4=microwave, 5=baseball glove, 6=donut. Answer with a single integer 1 to 6 and
nothing else.
\par}

\paragraph{Color ladder.} Templates, with the slots the harness fills in
braces. \{where\} is a position word such as bottom-center; \{options\} is the
$K$ calibrated words in the trial's order; \{which\} is ``Which of these is it
closest to?'' on held-out angles; \{angle\} is the model's own calibrated angle
for each word (Table~\ref{tab:models}).

{\small\ttfamily\raggedright
\textbf{word.} In the array you memorized, the square at the \{where\} of the
image was one of these colors: \{options\}. \{which\} Answer with a single color
word and nothing else.

\textbf{label.} \ldots{} one of these colors: 1=\{option 1\}, 2=\{option 2\},
\ldots{} \{which\} Answer with a single integer 1 to \{K\} and nothing else.

\textbf{numeric.} Colors are described by a hue angle from 0 to 359 on a color
wheel where \{angle\} is \{word\}, \ldots{} In the array you memorized, the
square at the \{where\} of the image had one of these hues: 1=\{angle 1\},
2=\{angle 2\}, \ldots{} \{which\} Answer with a single integer 1 to \{K\} and
nothing else.

\textbf{numeric, bare.} As numeric without the legend sentence.

\textbf{numeric, glossed.} \ldots{} had one of these hues: 1=\{angle 1\}
(\{word 1\}), 2=\{angle 2\} (\{word 2\}), \ldots

\textbf{numeric, mispaired.} As glossed, with the words permuted so that no
angle carries its own word.
\par}

\paragraph{Option order.} For each trial the order of the alternatives is a
deterministic permutation seeded by a hash of the trial identifier, so it is
random across trials and identical across conventions for the same trial. The
mispaired gloss is a derangement of the same seed, a permutation with no fixed
point, so no option is ever correctly glossed by chance. Orders were not
re-sampled; the paired design removes order effects from every within-item
contrast, and the selection-bias literature (\S\ref{sec:related}) is the place
to look for their marginal size.

\paragraph{Gate sensitivity.} The gate admits a model whose control condition
is at least 20 points over chance. On the color task the models showing the
penalty clear it by $38.2$ to $67.7$ points, so the set of penalty models is the
same for every threshold from $0$ to $38.2$; only the denominator moves. LLaVA
(null contrast) leaves above $20.8$ and SmolVLM (null contrast) enters below
$12.9$. On the photographs the penalty models clear by $34.9$ to $83.4$ and the
set is unchanged for any threshold up to $34.9$; GPT-4o (null) leaves above
$27.9$. No conclusion in the paper depends on the value 20.

\paragraph{Instrument agreement.} For every local model the crossmodel logs
hold both readouts on every visible trial. First-token argmax and the parsed
generation agree on $100.0\%$ of trials for Qwen2.5-VL-7B and InternVL3-2B on all
three rungs ($n=71$ each), on $100.0\%$ for LLaVA on the word rung ($n=48$), and
on $97.2\%$ for SmolVLM on the word rung ($n=71$; two trials). SmolVLM fails the
gate and contributes no contrast.

\section{Additional tables}
\label{app:tables}

\begin{table}[h]
\centering
\caption{Calibrated color vocabularies. Anchors are wheel angles at the center of
a run the model named unanimously across three renders.}
\label{tab:models}
\small
\begin{tabular}{ll}
\toprule
model & stable anchors (angle, name) \\
\midrule
Qwen3-VL-4B   & 5 pink, 50 orange, 95 yellow, 145 green, 245 blue, 310 purple \\
Qwen2.5-VL-7B & 55 orange, 100 yellow, 140 green, 185 teal, 245 blue, 340 pink \\
InternVL3-2B  & 30 pink, 100 yellow, 145 green, 205 cyan, 255 blue, 310 purple \\
SmolVLM-2B    & 30 red, 105 yellow, 145 green, 205 cyan, 260 blue, 315 purple \\
LLaVA-1.5-7B  & 40 pink, 145 green, 235 blue, 310 purple \\
\bottomrule
\end{tabular}
\end{table}

\begin{table}[h]
\centering
\caption{COCO \textsc{locate}, chance $11.1\%$, same 200 photographs. Gap is
cell name minus pixel, paired per image. Gemini's gap is $-15.5$ counting
unparseable answers as wrong and $-11.1$ on parseable answers only
($p=1.0\times10^{-4}$); the pixel prompt produces $5.5\%$ unparseable output
against $0.0\%$ for cell names. Frontier models as served on 20 September 2026.}
\label{tab:fivecoco}
\begin{tabular}{lrrrrr}
\toprule
model & cell name & normalized & pixel & gap & $p$ \\
\midrule
Qwen3-VL-4B    & 68.5 & 59.5 & 20.0 & $-48.5$ & $1.2\times10^{-21}$ \\
Qwen3-VL-4B, 8-bit & 76.0 & 68.5 & 24.0 & $-52.0$ & $5.0\times10^{-24}$ \\
InternVL3-2B   & 46.0 & 19.0 & 13.5 & $-32.5$ & $4.2\times10^{-13}$ \\
Qwen2.5-VL-7B  & 68.0 &  5.0 & 59.0 &  $-9.0$ & $0.08$ \\
Gemini (API)   & 94.5 & 94.0 & 79.0 & $-15.5$ & $3.4\times10^{-7}$ \\
GPT-4o (API)   & 39.0 & 22.0 & 34.0 &  $-5.0$ & $0.24$ \\
\bottomrule
\end{tabular}
\end{table}

\begin{table}[h]
\centering
\caption{Robustness of the Qwen3-VL-4B cell-name minus pixel gap on COCO
\textsc{locate}, by tercile of object size (fraction of image area) and of the
target center's distance from the nearest cell boundary (in cell widths). Exact
McNemar, $n$ images per slice.}
\label{tab:slices}
\begin{tabular}{llrrrr}
\toprule
slice & & cell name & pixel & gap & $p$ \\
\midrule
object size & small ($<0.026$)   & 66.7 & 18.2 & $-48.5$ & $4.4\times10^{-7}$ \\
            & medium             & 77.3 & 24.2 & $-53.0$ & $1.0\times10^{-8}$ \\
            & large ($>0.113$)   & 61.8 & 17.6 & $-44.1$ & $6.9\times10^{-8}$ \\
\midrule
boundary distance & near edge ($<0.30$)  & 62.1 & 25.8 & $-36.4$ & $7.0\times10^{-5}$ \\
                  & middling             & 72.7 & 13.6 & $-59.1$ & $2.2\times10^{-10}$ \\
                  & well inside ($>0.36$)& 70.6 & 20.6 & $-50.0$ & $5.4\times10^{-9}$ \\
\midrule
category & excluding \emph{person} & --- & --- & $-47.1$ & $3.3\times10^{-18}$ \\
\bottomrule
\end{tabular}
\end{table}

\begin{table}[h]
\centering
\caption{Synthetic spatial arrays, Qwen3-VL-4B, 120 trials, four colored squares
in distinct cells of a $3\times3$ grid. \textsc{locate}: 9 options, chance
$11.1\%$; \textsc{identify}: 6 options, chance $16.7\%$. Visible / cached. $p$ is
exact McNemar against cell names.}
\label{tab:synthspace}
\begin{tabular}{llrrr}
\toprule
direction & options described as & visible & cached & $p$ (visible / cached) \\
\midrule
\textsc{locate}   & cell name        & 70.8 & 68.3 & --- \\
                  & normalized 0--999 & 58.3 & 48.3 & $0.05$ / $1.2\times10^{-3}$ \\
                  & absolute pixels  & 42.5 & 35.8 & $8.2\times10^{-6}$ / $1.5\times10^{-7}$ \\
\midrule
\textsc{identify} & cell name        & 96.7 & 95.8 & --- \\
                  & normalized 0--999 & 90.8 & 90.0 & $0.12$ / $0.09$ \\
                  & absolute pixels  & 76.7 & 77.5 & $8.4\times10^{-6}$ / $2.7\times10^{-5}$ \\
\bottomrule
\end{tabular}
\end{table}

\end{document}